\documentclass[conference]{IEEEtran}
\IEEEoverridecommandlockouts
\usepackage{cite}
\usepackage{amsmath,amssymb,amsfonts,extarrows}
\usepackage{algorithm,algorithmic}
\usepackage{graphicx,subfig}
\usepackage{textcomp}
\usepackage{xcolor}
\def\BibTeX{{\rm B\kern-.05em{\sc i\kern-.025em b}\kern-.08em
    T\kern-.1667em\lower.7ex\hbox{E}\kern-.125emX}}
\begin{document}

\title{Exploitation and Exploration Analysis of Elitist Evolutionary Algorithms: A Case Study%*\\
\thanks{The second author was supported by EPSRC under Grant No. EP/I009809/1.}
}

\author{\IEEEauthorblockN{Yu Chen}
\IEEEauthorblockA{\textit{School of Science} \\
\textit{Wuhan University of Technology}\\
Wuhan, China \\
ychen@whut.edu.cn}
\and
\IEEEauthorblockN{Jun He}
\IEEEauthorblockA{\textit{School of Science and Technology} \\
\textit{Nottingham Trent University}\\
Nottingham, UK \\
jun.he@ntu.ac.uk}
}

\maketitle

\begin{abstract}
Known as two cornerstones of problem solving by search, exploitation and exploration are extensively discussed for implementation and application of evolutionary algorithms (EAs). However, only a few researches focus on evaluation and theoretical estimation of  exploitation and exploration. Considering that exploitation and exploration are two issues regarding global search and local search, this paper proposes to evaluate them via the success probability and the one-step improvement rate computed in different domains of integration. Then, case studies are performed by analyzing performances of (1+1) random univariate search and (1+1) evolutionary programming on the sphere function and the cheating problem. By rigorous theoretical analysis, we demonstrate that both exploitation and exploration of the investigated elitist EAs degenerate exponentially with the problem dimension $n$. Meanwhile, it is also shown that maximization of exploitation and exploration can be achieved by setting an appropriate value for the standard deviation $\sigma$ of Gaussian mutation, which is positively related to the distance from the present solution to the center of the promising region.
\end{abstract}


\section{Introduction}
An efficient evolutionary algorithm (EA) locates the absorbing basin of global optimal solution and refine its population to get high-precision solutions, which is widely known to be achieved by striking a balance between exploitation and exploration. Since Eiben and Schippers provided an early discussion on evolutionary exploitation and exploration in 1998~\cite{eiben1998evolutionary}, a lot of researches have been performed to discuss achievement and control of exploitation and exploration~\cite{vcrepinvsek2013exploration}, which results in emergence of a large amount of researches on implementation and application of EAs.

As presented in ~\cite{vcrepinvsek2013exploration}, exploitation and exploration are the two cornerstones of problem solving by search. However, delimitation of the two cornerstones is difficult and as yet unachievable. Since exploitation and exploration  are the processes of visiting entirely new regions and those regions within
the neighborhood of previously visited points, one common belief is that EAs should start with focus on exploration and then gradually change it onto exploitation. Thus, some of existing researches evaluated balances between exploitation and exploration by diversity metrics that degenerate as iteration continues ~\cite{alba2005exploration,inoue2015analyzing}.
It is well-known that performances of EAs are significantly influenced by landscapes of problems to be solved, and so, exploitation and exploration are related to not only population diversity but also fitness of individuals. Turkey and Poli~\cite{turkey2014model} defined an exploitation indicator that is an entropy-based measure assessing the dependency on fitness distributions of different features of population dynamics. More intuitively, Liu \emph{et al.}~\cite{liu2013parameter} defined a exploration metric as the percentage of nodes obtained by exploration over all nodes in all ancestry trees, and define the exploitation metric as 1 minus the exploration metric. The common viewpoint in these researches is that exploration exactly contradicts with exploitation.

Although excellent performances of EAs is achieved by balancing exploitation and exploration, it does not means that exploitation and exploration are necessarily conflicting. Since exploitation and exploration are two processes of visiting different part of the feasible region, they could be alternatively validated by the random mechanism of EAs. Thus, quantification of exploration/exploitation ability differs from implementing frequency of exploration/exploitation. In this paper, we would analyze exploration/exploitation of EAs by two metrics: the success probability and the expected ability of exploring/exploiting the feasible region. By computing the definition integrals in different regions, they can be employed to evaluate either exploitation or exploration. Then, exploitation and exploration analysis is performed based on estimation of these metrics.

Rest of this paper is organized as follows. Section \ref{PreSec} presents some preliminaries for further study of this paper; Then, case studies of exploitation analysis and exploration analysis are performed in Sections \ref{ExploitSec} and \ref{ExploreSec}, respectively. Finally, Section \ref{ConSec} concludes this paper.

\section{Preliminary}\label{PreSec}
In this paper, we only investigate how exploitation and exploration ability varies from the recombination operation and the problem dimension $n$. Thus, performance of EAs is studied by considering elitist single-individual EAs, where a candidate solution is accepted if and only if it is not worse than the present solution. Two popular instances of elitist single-individual EAs, for the binary-coded case, are the random local search (RLS) and the (1+1) evolutionary algorithm ((1+1)EA), where candidate solutions are generated by one-bit mutation and bitwise mutation, respectively.

Note that RLS and (1+1)EA are a local search algorithm and a global search algorithm, respectively. For the continuous problems studied in this paper, we investigate two continuous variants, named as the (1+1) random univariate search ((1+1)RUS) and the (1+1) evolutionary programming ((1+1)EP), respectively. The (1+1)RUS described by Algorithm \ref{RUS} performing Gaussian mutation on a randomly selected decision variables, is thus a local search algorithm; the (1+1)EP illustrated in Algorithm \ref{EP} employs simultaneous Gaussian mutation on all decision variables, which implies that it is a global one.

\begin{algorithm}[ht]
\caption{(1+1) Random Univariate Search}
\label{RUS}
\begin{algorithmic}[1]
\STATE  $t \leftarrow 0$;
\STATE  initialize a solution $\mathbf{x}_{0} =(x_1, \cdots, x_d) $ ;
\WHILE{the maximal number of generations is not reached}
\STATE   choose one index $j \in \{1, \cdots, d\}$ at random, and generate a new solution by $\mathbf{y}_t= \mathbf{x}_{t}+\mathbf{z}_t$  where $\mathbf{z}_t=(z_1, \cdots, z_d)$,  $z_{j}\sim \mathcal{N}(0,\sigma_j)$ is a Gaussian random variable and $z_{i}=0$ for other $i\neq j$; if $\mathbf{y}_t$ is out of the definition domain $\mathcal{D}$, let $\mathbf{y}_t=\mathbf{x}_t$;
\STATE  select the best one from $\mathbf{y}_t$ and $ \mathbf{x}_t $ as $\mathbf{x}_{t+1}$;
\STATE $t\leftarrow t+1$;
\ENDWHILE
\end{algorithmic}
\end{algorithm}

\begin{algorithm}[ht]
\caption{(1+1) Evolutionary Programming}
\label{EP}
\begin{algorithmic}[1]
\STATE  generation counter $t \leftarrow 0$;
\STATE  initialize an individual  $\mathbf{x}_{0}$;
\WHILE{$t$ is less than the maximal number of generations}
\STATE   generate a new individual by  Gaussian mutation  $\mathbf{y}_{t} = \mathbf{x}_{t}+\mathbf{z}_t$ where $\mathbf{z}_t$ obeys a Gaussian probability distribution; if $\mathbf{y}_t$ is beyond the definition domain $\mathcal{D}$, let $\mathbf{y}_t=\mathbf{x}_t$;
\STATE select the best one from $\mathbf{y}_t$ and $ \mathbf{x}_t $ as $\mathbf{x}_{t+1}$;
\STATE $t\leftarrow t+1$;
\ENDWHILE
\end{algorithmic}
\end{algorithm}

To perform exploitation/exploration analysis of EAs, we consider the continuous minimization problem
\begin{align}%\label{Sphere}
  \min\quad f(\mathbf x),\quad \mathbf{x}=(x_1,\dots,x_n)\in\mathbb R^n.
\end{align}
Denote the present solution $\mathbf{x}=(x_1,\dots, x_n)$ with $C=f(\mathbf x)$. For elitist EAs, only generated candidate solutions with smaller function values can be accepted. Denote the promising region as
\begin{align}\label{ProReg}
\mathcal{G}(C)=\{\mathbf{y}=(y_1,\dots,y_n) |f(\mathbf y)\le C)\}.
\end{align}
The success probability defined as
\begin{equation}\label{PC}
P(C)=\Pr\{\mathbf{y}\in\mathcal{G}(C)|\mathbf{x}\}=\int_{\mathbf y\in\mathcal G(C)}d\mathbf P,
\end{equation}
is a metric measuring the ability to avoid stagnation; the one-step improvement rate, defined as
\begin{equation}\label{IRC}
IR(C)=\mathbb E\{\mathbf{y}\in\mathcal{G}(C)|\mathbf{x}\}/C=\int_{\mathbf y\in\mathcal G(C)}(C-f(\mathbf y))d\mathbf P/C,
\end{equation}
is a metric of convergence speed.

Note that both (\ref{PC}) and (\ref{IRC}) evaluate the ability of EAs to search the region $\mathcal{G}(C)$. Since exploration visits entirely new regions and exploitation visits the neighborhood of previously visited points, we can set different domains of integration in (\ref{PC}) and (\ref{IRC}) to evaluate exploitation and exploration of EAs. For exploration analysis, the domain of integration is taken as promising subregions that are away from the present solution $\mathbf{x}$; however, we take the promising subregion adjacent to $\mathbf{x}$ as the domain of integration for exploitation analysis.

\section{Exploitation Analysis}\label{ExploitSec}
To perform exploitation analysis, we consider minimization of the sphere function
\begin{align}\label{Sphere}
  \min\quad f_{sph}(\mathbf x)=\sum_{i=1}^{n}x_i^2,\quad \mathbf{x}=(x_1,\dots,x_n)\in\mathbb R^n.
\end{align}
Given $C>0$, the present solution is $\mathbf{x}=(x_1,\dots,x_n)$ with $$f_{sph}(\mathbf x)=\sum_{i=1}^{n}x_i^2=C.$$
The promising region is a sphere adjacent to $\mathbf x$:
\begin{equation}\label{PR_S}
  \mathcal{G}_{sph}(C)=\{\mathbf{y}\in\mathbb{R}^n|\|\mathbf{y}\|^2\le C\}.
\end{equation}
Thus, $\mathcal{G}_{sph}(C)$ is the region exploitation visits, and we will investigate the probability (\ref{PC}) and the improvement rate (\ref{IRC}) for the promising region $\mathcal{G}_{sph}(C)$.

\subsection{Estimation of the Success Probability}\label{S1}
Note that (1+1)RUS performs the same as (1+1)EP. So, we first estimate for the 1-D sphere function the probability to hit the promising region, and then, investigate the (1+1)RUS and the (1+1)EP for $n>1$, respectively.

\subsubsection{$n=1$}
Denote the present solution as $x$, and the standard deviation of Gaussian mutation as $\sigma$. Then,
\begin{align}
  &\Pr\{y\in\mathcal{G}_{sph}(x^2)|x\}=\frac{1}{\sqrt{2\pi}\sigma}\int_{-x}^{x}e^{-\frac{(y-x)^2}{2\sigma^2}}dy\nonumber\\
  =&\Phi\left(\frac{2x}{\sigma}\right)-\Phi(0),\label{ProbSph1D}
\end{align}
where $\Phi(\cdot)$ is the cumulative distribution function (CDF) of standard Gaussian distribution. It is obvious that $\Pr\{\mathbf y\in\mathcal{G}_{sph}(C)\}$ increase as $\sigma$ decreases, and its supremum is $\frac{1}{2}$.

When $x=\sqrt{C}$, we know that the probability to hit the promising region
\begin{equation}\label{Prob_U_E_1}
  P_{sph}^{(1)}(C)=\Phi\left(\frac{2\sqrt{C}}{\sigma}\right)-\frac{1}{2}.
\end{equation}

\subsubsection{(1+1)RUS} Denote the present solution as $\mathbf x=(x_1,\dots,x_n)$ with $\sum_{i=1}^nx_i^2=C$. Without loss of generality, suppose that $x_i\ge 0,i=1,\dots,n$. When $x_i$ is selected with probability $\frac{1}{n}$ to be mutated, other $n-1$ variables keep unchanged. Then, a better solution $\mathbf y=(x_1,\dots,x_{i-1},y_i,x_{i+1},\dots,x_n)$ is generated if and only if $|y_i|\le x_i$. Thus, The success probability of (1+1)RUS to hit the promising region is
\begin{align}%\label{}
  &P_{sph}^{R}(C)=\Pr\{\mathbf y\in\mathcal{G}_{sph}(C)|\|\mathbf{x}\|^2=C\}\nonumber\\
  =&\sum_{i=1}^n\frac{1}{n}\Pr\{|y_i|\le x_i|\|\mathbf{x}\|^2=C\}\nonumber\\
  =&\frac{1}{n}\sum_{i=1}^n\frac{1}{\sqrt{2\pi}\sigma}\int_{-x_i}^{x_i}e^{\frac{(y_i-x_i)^2}{2\sigma^2}}dy_1\dots dy_n\nonumber\\
  =&\frac{1}{n}\sum_{i=1}^n\left[\Phi\left(\frac{2x_i}{\sigma}\right)-\frac{1}{2}\right]=\frac{1}{n}\sum_{i=1}^nP_{sph}^{(1)}(x_i^2).
\end{align}

Denote $x_{max}=\max_{i=1,\dots,n}x_i$. Then, $x_{max}\ge\sqrt{C/n}$, and we know that
\begin{align}\label{B1}
  \textstyle P_{sph}^{R}(C) \ge\frac{1}{n}\left[\Phi\left(\frac{2x_{max}}{\sigma}\right)-\frac{1}{2}\right] \ge\frac{1}{n}\left[\Phi\left(\frac{2\sqrt{C}}{\sigma\sqrt{n}}\right)-\frac{1}{2}\right].
\end{align}
Meanwhile, by defining $\Psi(z)=\int_0^z\frac{e^{-\frac{z}{2}}}{\sqrt{z}}dz,$ we know that
\begin{align*}
  & P_{sph}^{(1)}(C)=\frac{1}{\sqrt{2\pi}}\int_{0}^{\frac{2x_i}{\sigma}}e^{\frac{y^2}{2}}dy \\ =&\frac{1}{\sqrt{2\pi}}\int_{0}^{\frac{4x_i^2}{\sigma^2}}\frac{e^{-\frac{z}{2}}}{\sqrt{z}}dz=\frac{1}{\sqrt{2\pi}}\Psi\left(\frac{4x_i^2}{\sigma^2}\right).
\end{align*}
Consider the second-order derivative
\begin{align*}
  \Psi''(z)=-\frac{1}{2}e^{-\frac{z}{2}}\left(z^{-\frac{1}{2}}+z^{-\frac{3}{2}}\right),
\end{align*}
which is negative when $z>0$. That is to say, $\Psi(z)$ is concave, and
\begin{align}
  &P_{sph}^{R}(C)=\frac{1}{n}\sum_{i=1}^n\Psi\left(\frac{4x_i^2}{\sigma^2}\right) \le\Psi\left(\frac{1}{n}\sum_{i=1}^n\frac{4x_i^2}{\sigma^2}\right)\nonumber\\
  =&\Psi\left(\frac{4C}{\sigma^2n}\right)=\Phi\left(\frac{2\sqrt{C}}{\sigma\sqrt{n}}\right)-\frac{1}{2}.\label{U1}
\end{align}
Combining (\ref{B1}) and (\ref{U1}) we know that
\begin{equation}\label{Pro_RUS}
 \textstyle\frac{1}{n}\left[\Phi\left(\frac{2\sqrt{C}}{\sigma\sqrt{n}}\right)-\frac{1}{2}\right] \le P_{sph}^{R}(C) \le\Phi\left(\frac{2\sqrt{C}}{\sigma\sqrt{n}}\right)-\frac{1}{2}.
\end{equation}

\subsubsection{(1+1)EP}
According to symmetry of the probability dense function (PDF) of Gaussian distribution, we can suppose $\mathbf{x}=(\sqrt{C},0,\dots,0)$ without loss of generality. Then
\begin{align}
  &P_{sph}^{E}(C)=\Pr\{\mathbf y\in\mathcal{G}_{sph}(C)|\|\mathbf{x}\|^2=C\}\nonumber\\
  =&\frac{1}{(\sqrt{2\pi}\sigma)^{n}}\int_{\mathbf y\in\mathcal{G}_U}e^{\frac{(y_1-\sqrt{C})^2+\sum_{i=2}^ny_i^2}{2\sigma^2}}dy_1\dots dy_n\nonumber\\
  =&\frac{1}{(\sqrt{2\pi}\sigma)^{n}}\int_{-\sqrt{C}}^{\sqrt{C}}e^{-\frac{(y_1-\sqrt{C})^2}{2\sigma^2}}dy_1\nonumber\\
  &\int_{\sum_{i=2}^ny_i^2\le\sqrt{C-y_1^2}}e^{-\frac{\sum_{i=2}^ny_i^2}{2\sigma^2}}dy_2\dots dy_n\nonumber\\
  \le&\textstyle\frac{1}{(\sqrt{2\pi})^{n}}\int_{-\sqrt{C}}^{\sqrt{C}}e^{-\frac{(y_1-\sqrt{C})^2}{2\sigma^2}}dy_1 \left(\int_{-\sqrt{C}}^{\sqrt{C}}e^{-\frac{y^2}{2\sigma^2}}dy\right)^{n-1}\nonumber\\
  =&\textstyle\left(\Phi\left(\frac{2\sqrt{C}}{\sigma}\right)- \frac{1}{2}\right)\left(\Phi\left(\frac{\sqrt{C}}{\sigma}\right)-\Phi\left(-\frac{\sqrt{C}}{\sigma}\right)\right)^{n-1}. \label{ProbSphnDU}
\end{align}

A lower bound of $P_{sph}^{E}(C)$ can be obtained by supposing without loss of generality thatp $\mathbf{x}=(x,\dots,x)=(\sqrt{\frac{C}{n}},\dots,\sqrt{\frac{C}{n}})$ . Then, the hyperrectangle $\mathcal{G}_B=\left[-\sqrt{\frac{C}{n}},\sqrt{\frac{C}{n}}\right]^n$ is included in $\mathcal{G}_{sph}(C)$, which implies that
$$P_{sph}^E(C)\ge \Pr\{\mathbf y\in\mathcal{G}_B|\| x\|^2=C\}.$$
By (\ref{ProbSph1D}) we know that
\begin{align*}%\label{ProbB_nd}
  \Pr\{\mathbf y\in\mathcal{G}_B|\| x\|^2=C\}=\left(\Phi\left(\frac{2\sqrt{C}}{\sigma\sqrt{n}}\right)-\frac{1}{2}\right)^n,
\end{align*}
and thus,
\begin{align}\label{ProbSphnDB}
  P_{sph}^E(C)\ge\left(\Phi\left(\frac{2\sqrt{C}}{\sigma\sqrt{n}}\right)-\frac{1}{2}\right)^n.
\end{align}

\begin{figure*}[!t]
\centering
\subfloat[Numerical simulation of $P_E(C)$]{\includegraphics[width=2in]{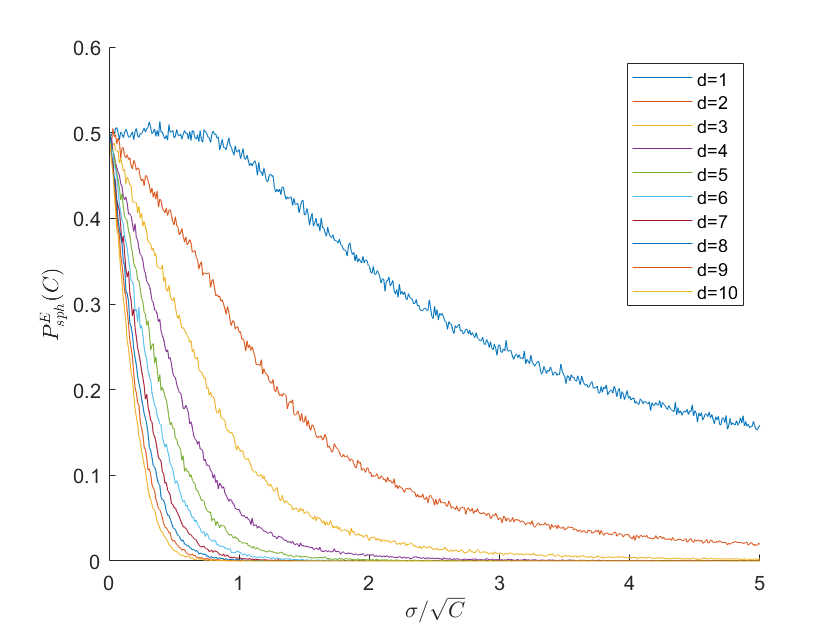}
\label{fig_first_case}}
\hfil
\subfloat[The lower bound $P^l_E(C)$]{\includegraphics[width=2in]{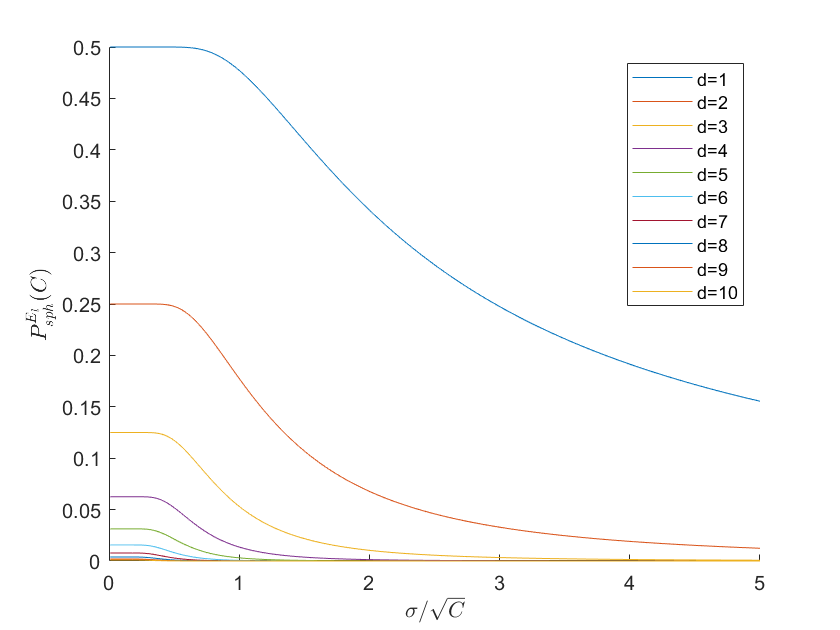}
\label{fig_second_case}}
\hfil
\subfloat[The upper bound $P^u_E(C)$]{\includegraphics[width=2in]{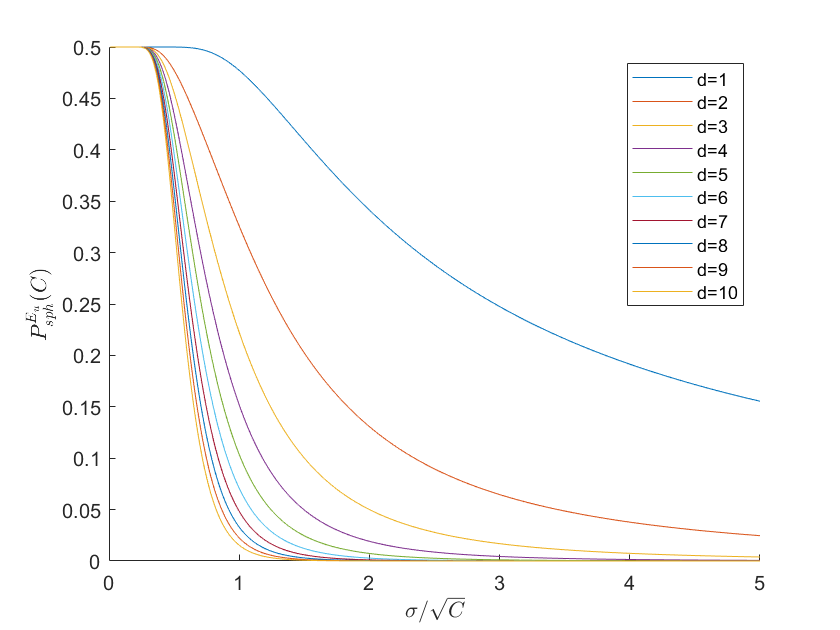}
\label{fig_third_case}}
\caption{Trend plots of the probability $P_{sph}^E(C)$, its lower bound $P^{E_b}_{sph}(C)$ and its upper bound $P^{E_u}_{sph}(C)$ with increase of the ratio $\sigma/\sqrt{C}$.}
\label{PR_sigma}
\end{figure*}

Denote the upper bound and the lower bound presented in (\ref{ProbSphnDU}) and (\ref{ProbSphnDB}) as $P^{E_u}_{sph}(C)$ and $P^{E_l}_{sph}(C)$, respectively. It is obvious that both of them decrease exponentially with problem dimension $n$, and increase as $\sigma/\sqrt{C}$ decreases. While $n=1$, (\ref{ProbSphnDU}) and (\ref{ProbSphnDB}) presents an precise expression of $P_{sph}^E(C)$; however, just as illustrated in Fig. \ref{PR_sigma}, curves of the upper bounds $P^{E_u}_{sph}(C)$ (Fig. \ref{fig_third_case}) can better accommodate the numerical simulation results of $P_{sph}^{E}(C)$, which indicates that (\ref{ProbSphnDU}) is an appropriate estimation of $P_{sph}^{E}(C)$. That is,
\begin{equation}\label{P_E}
  P_E(C)=\mathcal{O}(a^{n-1}),\quad 0<a<1.
\end{equation}

%\begin{figure}[htbp]
%\centerline{\includegraphics[height=5cm]{PR.png}}
%\caption{Example of a figure caption.}
%\label{PR_sigma}
%\end{figure}
%From (\ref{ProbSphnDU}) and (\ref{ProbSphnDB}) we know that the probability to generate bett%er solution degenerates exponentially with problem dimension $n$. A small $\sigma$ can slow down its degenerating speed with increase of $n$.

\subsection{Estimation of the  One-step Improvement Rate}
As discussed in section \ref{S1}, the probability $P_{sph}^{E}(C)$ increase as $\sigma$ decrease, which implies that we can set it as small as possible to avoid stagnation of iteration. However, it does not imply that a small value of $\sigma$ can lead to fast convergence of EAs. To answer this question in a rigorous way, we investigate the connection between convergence speed and influential factors by estimating the one-step improvement rate of function value.

\subsubsection{$n=1$}
For the present solution $x$, the elitist selection contributes to a one-step expected improvement of fitness
\begin{align}\label{FitSph1D}
  &\mathbb{E}[f(x)-f(y)|x]=\frac{1}{\sqrt{2\pi}\sigma}\int_{-x}^{x}(x^2-y^2)e^{-\frac{(y-x)^2}{2\sigma^2}}dy\nonumber\\
  =&x^2\left(\Phi\left(\frac{2x}{\sigma}\right)-\frac{1}{2}\right)-\frac{1}{\sqrt{2\pi}}\int_{-\frac{2x}{\sigma}}^0(\sigma z+x)^2e^{-\frac{z^2}{2}}dz\nonumber\\
  =&x^2\left(\Phi\left(\frac{2x}{\sigma}\right)-\frac{1}{2}\right)-\frac{1}{\sqrt{2\pi}}\left[\int_{-\frac{2x}{\sigma}}^0\sigma^2 z^2e^{-\frac{z^2}{2}}dz\right.\nonumber\\ &\left. +\int_{-\frac{2x}{\sigma}}^0x^2e^{-\frac{z^2}{2}}dz+\int_{-\frac{2x}{\sigma}}^0 2\sigma xze^{-\frac{z^2}{2}}dz\right].
\end{align}
Denote
\begin{align}
  I_1(a,b) &=\frac{1}{\sqrt{2\pi}}\int_{a}^{b} z^2e^{-\frac{z^2}{2}}dz=-\frac{1}{\sqrt{2\pi}}\left.ze^{-\frac{z^2}{2}}\right|_{a}^{b}\nonumber\\
  &+\left(\Phi(b)-\Phi(a)\right), \label{I_1}
\end{align}
\begin{align}
  I_2(a,b) &=\frac{1}{\sqrt{2\pi}}\int_{a}^{b} e^{-\frac{z^2}{2}}dz=\Phi(b)-\Phi(a), \label{I_2}
\end{align}
and
\begin{align}
  I_3(a,b) &=\frac{1}{\sqrt{2\pi}}\int_{a}^{b} ze^{-\frac{z^2}{2}}dx=-\frac{1}{\sqrt{2\pi}}\left.e^{-\frac{x^2}{2}}\right|_{a}^{b}.\label{I_3}
\end{align}
(\ref{FitSph1D}), (\ref{I_1}), (\ref{I_2}) and (\ref{I_3}) imply that
\begin{align*}
  &\mathbb{E}\left[f(x)-f(y)|x\right] =\textstyle x^2\left(\Phi\left(\frac{2x}{\sigma}\right)-\frac{1}{2}\right)\\ &\textstyle -\left[\sigma^2I_1\left(0,-\frac{2x}{\sigma}\right)+x^2I_2\left(0,-\frac{2x}{\sigma}\right)+2\sigma xI_3\left(0,-\frac{2x}{\sigma}\right)\right]\\
  =&\textstyle\frac{2\sigma x}{\sqrt{2\pi}}-\sigma^2\left(\frac{1}{2}-\Phi\left(-\frac{2x}{\sigma}\right)\right).
\end{align*}
Let the present solution be $x=\sqrt{C}$. The conditional improvement rate is
\begin{align}
  &IR^{(1)}_{sph}(C)=\mathbb{E}\left[f(x)-f(y)|x=\sqrt{C}\right]/C\nonumber\\
   =&\textstyle\frac{2\sigma }{\sqrt{2\pi C}}-\frac{\sigma^2}{C}\left(\frac{1}{2}-\Phi\left(-\frac{2\sqrt{C}}{\sigma}\right)\right).\label{OneStepImpRate1}
\end{align}

\begin{figure}[htbp]
\centerline{\includegraphics[height=5cm]{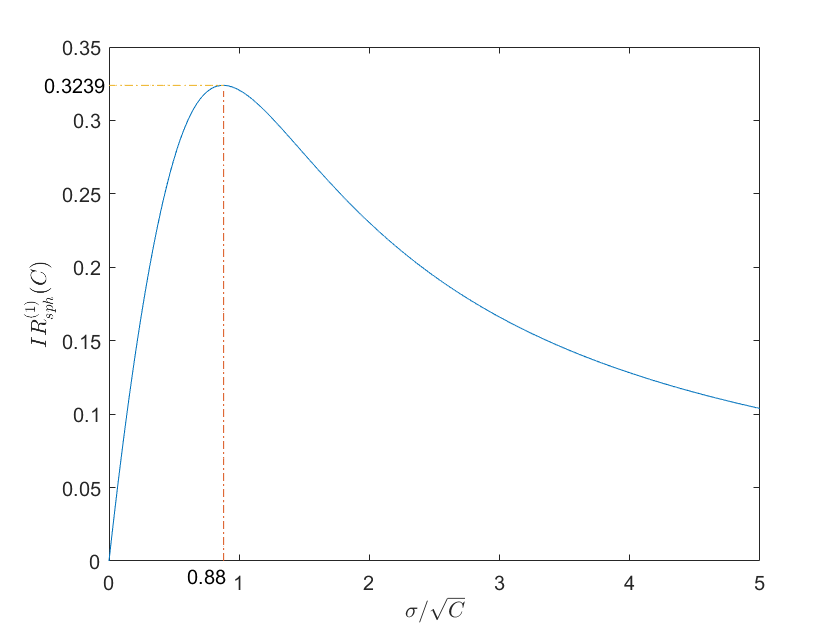}}
\caption{Trend plot of the one-step improvement rate of fitness for $n=1$.}
\label{CR}
\end{figure}

Theoretical analysis on monotonicity of (\ref{OneStepImpRate1}) is tedious, and we investigate it by numerical simulation. Fig. \ref{CR} illustrates that the one-step improvement rate first increases as $\sigma$ increase, then, reaches its maximum  value 0.3239 when $\frac{\sigma}{\sqrt{C}}$ is around $0.88$, and monotonously tends to zero as $\sigma\to 0$.

\subsubsection{(1+1)RUS}
Since (1+1)RUS perform Gaussian mutation on a randomly selected variable, the expected one-step improvement truncated by elitist selection is
\begin{align}%\label{}
  &\mathbb{E}[f(\mathbf{x})-f(\mathbf{y})|\mathbf{x}]=\sum_{i=1}^n\frac{1}{n}\mathbb{E}[f(x_i)-f(y)|x_i]\nonumber\\
  =&\frac{1}{n} \sum_{i=1}^n\left[\frac{2\sigma x_i}{\sqrt{2\pi}}-\sigma^2\left(\Phi\left(\frac{2x_i}{\sigma}\right)-\frac{1}{2}\right)\right].
\end{align}
For the case that $\|\mathbf{x}\|^2=\sum_{i=1}^nx_i^2=C$, the one-step improvement rate of (1+1)RUS is
\begin{align}%\label{}
  &IR_{sph}^R(C)=\mathbb{E}[f(\mathbf{x})-f(\mathbf{y})|\|\mathbf{x}\|^2=C]/C \nonumber\\
  =&\frac{1}{nC} \sum_{i=1}^n\left[\frac{2\sigma x_i}{\sqrt{2\pi}}-\sigma^2\left(\Phi\left(\frac{2x_i}{\sigma}\right)-\frac{1}{2}\right)\right]\nonumber\\
  =&\frac{1}{n} \sum_{i=1}^n\frac{x_i^2}{C}\left[\frac{2\sigma }{\sqrt{2\pi}x_i}-\frac{\sigma^2}{x_i^2}\left(\Phi\left(\frac{2x_i}{\sigma}\right)-\frac{1}{2}\right)\right]\nonumber\\
  =&\frac{1}{n} \sum_{i=1}^n\frac{x_i^2}{C}IR^{(1)}_{sph}(x_i^2).
\end{align}
That is to say,
\begin{equation}\label{IR_C}
  IR_{sph}^R(C)=\Theta(n^{-1}).
\end{equation}
According to (\ref{OneStepImpRate1}) we know that the maximum of $IR^{(1)}_{sph}(x_i^2)$ is $0.88$. Thus, the optimal value of $IR_R(C)$ is $\frac{0.88}{n}$.

\subsubsection{(1+1)EP}
The Gaussian mutation performed by (1+1)EP is
\begin{align*}%\label{Gaussion}
  \mathbf y=\mathbf x+\mathcal{N}(0,\boldsymbol{\sigma}),
\end{align*}
where $\boldsymbol{\sigma}=(\sigma,\dots,\sigma)$. Then, thep one-step improvement of fitness is bounded from above by
  \begin{align*}%\label{FitSphnD}
  \mathbb{E}[f(\mathbf{x})-f(\mathbf{y})|\|\mathbf{x}\|^2=C]\le f(\mathbf{x})P_{sph}^E(C).
\end{align*}
By (\ref{ProbSphnDU}) we know that the one-step improvement rate of (1+1)EP is bounded from above by
\begin{align}\label{OneStepImpRaten}
  &IR_{sph}^E(C)=\mathbb{E}[f(\mathbf{x})-f(\mathbf{y})|\|\mathbf{x}\|^2=C]/C\le P_{sph}^E(C)\nonumber\\
 = &\textstyle \left(\Phi\left(\frac{2\sqrt{C}}{\sigma}\right)- \frac{1}{2}\right)\left(\Phi\left(\frac{\sqrt{C}}{\sigma}\right)-\Phi\left(-\frac{\sqrt{C}}{\sigma}\right)\right)^{n-1}.
\end{align}

Note that (\ref{OneStepImpRaten}) only provides a general bound for the one-step improvement rate, which indicates that $IR_{sph}^E(C)$ could not exceed $\frac{1}{2}$, no matter what value of $\sigma$ is employed. That is to say, the (1+1)EP could not be super-linearly convergent when it is employed minimizing the sphere function.

To obtain a lower bound of $IR_E(C)$, we suppose
$$\mathbf{x}=(x,\dots,x)=(\sqrt{\frac{C}{n}},\dots,\sqrt{\frac{C}{n}}).$$
Then,
\begin{align*}%\label{}
  &\mathbb{E}[f(\mathbf{x})-f(\mathbf{y})|\|\mathbf{x}\|^2=C]\\
  =&\textstyle\int_{\mathcal G_{sph}(C)}(C-\sum_{i=1}^ny_i^2)d\mathbf{P}\ge \int_{\mathcal G_B}(C-\sum_{i=1}^ny_i^2)d\mathbf{P}\nonumber\\
  =&\textstyle C\left(\Phi\left(\frac{2\sqrt{C}}{\sigma\sqrt{n}}\right)-\frac{1}{2}\right)^n-\int_{\mathcal G_B}\sum_{i=1}^ny_i^2d\mathbf{P}\nonumber\\
  =&\textstyle\left(\Phi\left(\frac{2\sqrt{C}}{\sigma\sqrt{n}}\right)-\frac{1}{2}\right)^{n-1}\left(\frac{2\sigma \sqrt{nC}}{\sqrt{2\pi C}}-n\sigma^2\left(\Phi\left(\frac{2\sqrt{C}}{\sigma\sqrt{n}}\right)-\frac{1}{2}\right)\right).
\end{align*}
Thus, it holds that
\begin{align}
  &\textstyle IR_{sph}^{E}(C) \ge \left(\Phi\left(\frac{2\sqrt{C}}{\sigma\sqrt{n}}\right)-\frac{1}{2}\right)^{n-1}\nonumber\\
  &\quad\quad\cdot\textstyle\left[\frac{2\sigma \sqrt{n}}{\sqrt{2\pi C}}-\frac{n\sigma^2}{C}\left(\frac{1}{2}-\Phi\left(-\frac{2\sqrt{C}}{\sigma\sqrt{n}}\right)\right)\right].\label{OneStepImpRaten1}
\end{align}
Denote
\begin{align*}
  I_4 &=\textstyle\left(\Phi\left(\frac{2\sqrt{C}}{\sigma\sqrt{n}}\right)-\frac{1}{2}\right)^{n-1},\\
  I_5 &=\textstyle\frac{2\sigma \sqrt{n}}{\sqrt{2\pi C}}-\frac{n\sigma^2}{C}\left(\frac{1}{2}-\Phi\left(-\frac{2\sqrt{C}}{\sigma\sqrt{n}}\right)\right).
\end{align*}
It is obvious that $I_3$ degenerates with increase of problem dimension $n$, and decreases as $\sigma$ increase. Similar to (\ref{OneStepImpRate1}), $I_4$ reaches its maximum 0.3239 value when $\sqrt{n}\sigma/\sqrt{C}$ is around $0.88$. That is, $I_4$ is maximized when $\sigma/\sqrt{C}=0.88/\sqrt{n}$.

\begin{figure*}[!t]
\centering
\subfloat[Numerical simulation of $IR_E(C)$]{\includegraphics[width=2in]{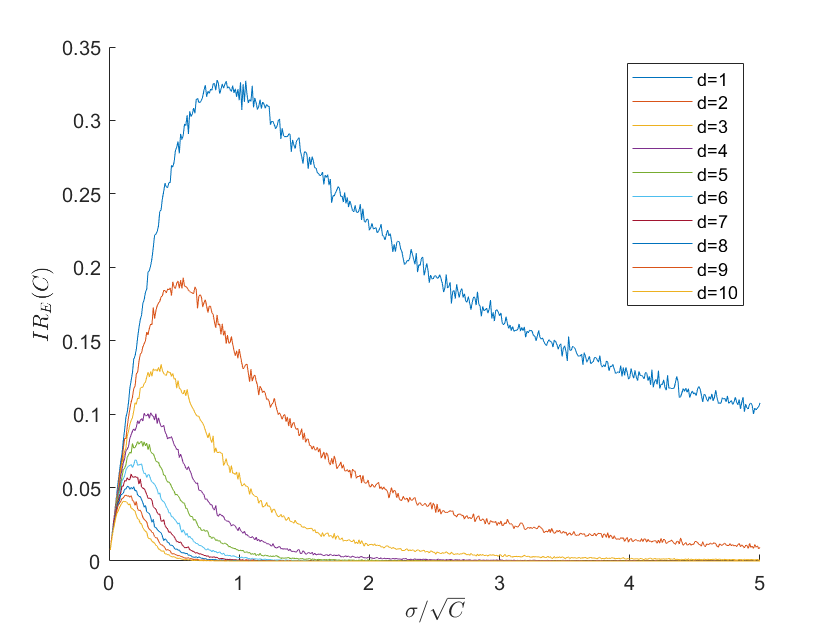}
\label{cfig_first_case}}
\hfil
\subfloat[The lower bound $IR^l_E(C)$]{\includegraphics[width=2in]{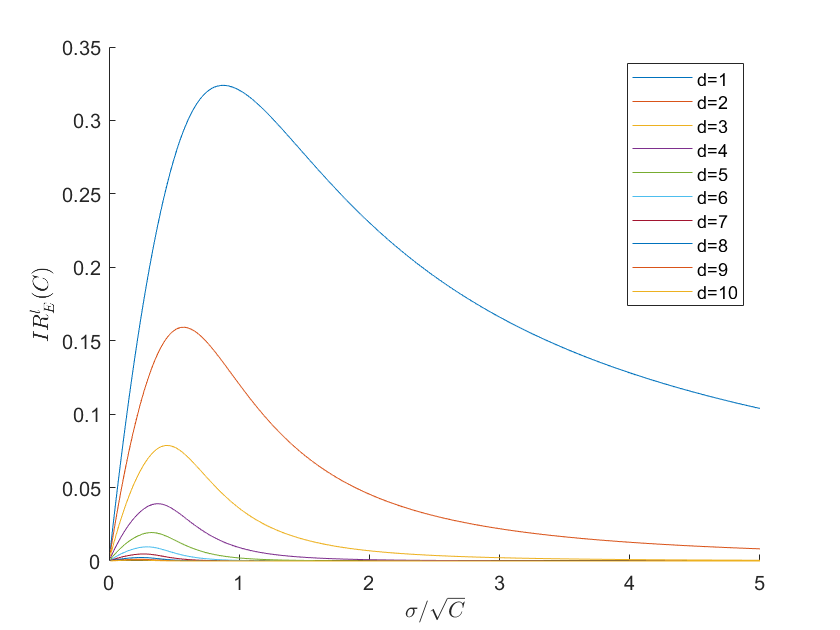}
\label{cfig_second_case}}

\caption{Trend plots of the improvement rate $IR_E(C)$ and its lower bound $P^b_E(C)$ with increase of the ratio $\sigma/\sqrt{C}$.}
\label{CR_sigma}
\end{figure*}

In conclusion, we know that when $n=1$, (\ref{OneStepImpRaten1}) degenerates to (\ref{OneStepImpRate1}), and the optimal setting of $\sigma$ is $\sigma_1^{*}\approx 0.88$; while $n>1$, the lower bound presented in (\ref{OneStepImpRaten1}) can reach its maximum value at some value $\sigma_n^{*}$ that is less than $0.88$ and decreases with $n$. Fig. \ref{CR_sigma} presents an illustrative comparison between $IR_{sph}^E(C)$ and its lower bound (\ref{OneStepImpRaten1}). Although values of (\ref{OneStepImpRaten1}) are significantly smaller than $IR_{sph}^E(C)$ when $n>1$, they exhibit similar profiles. From (\ref{OneStepImpRaten}) and (\ref{OneStepImpRaten1}) we can conclude that
\begin{equation}\label{IR_C1}
  IR_{sph}^E(C)=\Theta(a^{n-1}),\quad a\in(0,1).
\end{equation}

%a tighter upper bound could be obtained as follows.
%\begin{align}\label{FitSphnD1}
%  &\mathbb{E}[f(\mathbf{x}_0)-f(\mathbf{x})|\mathbf{x}_0]\nonumber\\
%  =&\int_{\mathcal G}f(\mathbf{x}_0)-f(\mathbf{x})dP \le\int_{\mathcal G_U}f(\mathbf{x}_0)-f(\mathbf{x})dP\nonumber\\
%  \le &f(\mathbf x_0)\left(\Phi\left(\frac{2x_0}{\sigma}\right)-\frac{1}{2}\right)^n-\prod_{i=1}^n\left[\frac{2\sigma x_i}{\sqrt{2\pi}}-\sigma^2\left(\frac{1}{2}-\Phi\left(-\frac{2x_i}{\sigma}\right)\right)\right]\nonumber\\
%  =&x_0^2\left\{\left(\Phi\left(\frac{2x_0}{\sigma}\right)-\frac{1}{2}\right)^n-\prod_{i=1}^n\left[\frac{2\sigma x_i}{\sqrt{2\pi}}-\sigma^2\left(\frac{1}{2}-\Phi\left(-\frac{2x_i}{\sigma}\right)\right)\right]\right\}.
%\end{align}

\section{Exploration Analysis}\label{ExploreSec}
To perform exploration analysis, we consider a continuous cheating problem
\begin{align}\label{Cheat}
  \min f_{cht}(\mathbf x)&=\left\{
  \begin{aligned}&\sum_{i=1}^nx_i^2,&&\mbox{if } \sum_{i=1}^nx_i^2\le M,\\ &(2M+1)-\sum_{i=1}^nx_i^2,&&\mbox{if } M<\sum_{i=1}^nx_i^2\le 2M,\end{aligned}\right.
\end{align}
where $\|\mathbf{x}\|=\sum_{i=1}^nx_i^2\le 2M$, $\mathbf{x}=(x_1,\dots,x_n)\in\mathbb R^n$.
Note that decision region of the cheating problem consists of two different sections: the ``absorbing region'' of the global optimal solution $\mathbf{x}^*=(0,\dots,0)$, denoted as $$\mathcal{G}_{cht}(M)=\{\mathbf{y}\in\mathbb{R}^n|\|\mathbf{y}\|\le M\},$$and the cheating region denoted as
 $$\overline{\mathcal{G}}_{cht}(M)=\{\mathbf{y}\in\mathbb{R}^n|M<\|\mathbf{y}\|\le 2M\}.$$

For the present solution $\mathbf{x}$ with $f_{cht}(\mathbf x)=C$, there are two cases to be distinguished.
\begin{enumerate}
  \item If $M<C\le M+1$, $\mathbf x$ is located in the cheating region $\overline{\mathcal{G}}_{cht}(M)$. For this case, $$\|\mathbf{x}\|=\sum_{i=1}^nx_i^2=2M+1-C;$$
  \item If $0<C\le M$, $\mathbf x$ could be located in either $\mathcal{G}_{cht}(M)$ or $\overline{\mathcal{G}}_{cht}(M)$. When $\mathbf x$ is located in $\overline{\mathcal{G}}_{cht}(M)$,
      $$\|\mathbf{x}\|=\sum_{i=1}^nx_i^2=2M+1-C;$$
      When $\mathbf x$ is located in ${\mathcal{G}}_{cht}(M)$,
      $$\|\mathbf{x}\|=\sum_{i=1}^nx_i^2=C;$$
\end{enumerate}

Due to the infinite search range of n-D Gaussian mutation, either exploration or exploitation could be performed during the iteration process.
\begin{enumerate}
\item When the present solution $\mathbf{x}$ is located in $\mathcal{G}_{cht}(M)$ ($\overline{\mathcal{G}}_{cht}(M)$), exploitation is performed if accepted solution $\mathbf{y}$ is also located in $\mathcal{G}_{cht}(M)$ ( $\overline{\mathcal{G}}_{cht}(M)$).
  \item When the present solution $\mathbf{x}$ located in $\mathcal{G}_{cht}(M)$ ($\overline{\mathcal{G}}_{cht}(M)$) is replaced by another solution $\mathbf{y}$ located in $\overline{\mathcal{G}}_{cht}(M)$ ( $\mathcal{G}_{cht}(M)$), exploration is performed. Exploration driving $\mathbf{x}\in\mathcal{G}_{cht}(M)$ to $\mathbf{y}\in\overline{\mathcal{G}}_{cht}(M)$ get a solution farther to the global optimal solution, which is called ``mistaken''; exploration driving $\mathbf{x}\in\overline{\mathcal{G}}_{cht}(M)$ to $\mathbf{y}\in\mathcal{G}_{cht}(M)$ get a solution nearer to the  global optimal solution, which is called ``right''.
\end{enumerate}

Since exploitation enumerated for the first case is the same as what we have discussed in Section \ref{ExploitSec}, we would not investigate it any more. Meanwhile, the ``mistaken'' exploration drives search in a false way, which does not indicate the ability of EAs to search for the global optimal solution, and so, we do not discuss it in this paper. Thus, in the following we only investigate the ``right'' exploration driving individuals jumping from $\overline{\mathcal{G}}_{cht}(M)$ to ${\mathcal{G}}_{cht}(M)$.
Then, for the present solution $\mathbf{x}=(x_1,\dots x_n)$ with $f_{cht}(\mathbf x)=C$, we have $$\|\mathbf{x}\|^2=\sum_{i=1}^nx_i^2=2M+1-C.$$ The probability to hit the promising region is
\begin{align*}
P_{cht}(C)=\left\{\begin{aligned}&\textstyle\int_{\|\mathbf y\|^2\le M}d\mathbf P, && \mbox{if } M\le C\le M+1,\\
&\textstyle\int_{\|\mathbf y\|^2\le C}d\mathbf P, &&\mbox{if }C< M, \end{aligned}\right.
\end{align*}
and the one-step improvement is
\begin{align*}
&IM_{cht}(C)\\
=&\left\{\begin{aligned}&\textstyle\int_{\|\mathbf y\|^2\le M}(C-\|\mathbf{y}\|^2)d\mathbf P, && \mbox{if } M\le C\le M+1,\\
&\textstyle\int_{\|\mathbf y\|^2\le C}(C-\|\mathbf{y}\|^2)d\mathbf P, &&\mbox{if }C< M. \end{aligned}\right.
\end{align*}

Similar to the exploitation analysis presented in Section \ref{ExploitSec}, we perform exploration analysis by estimating the success probability  and the one-step improvement rate to hit $\mathcal{G}_{cht}(C)$. For each estimation, we first investigate the case that $n=1$, for which (1+1)RUS performs the same as (1+1)EP. Then, (1+1)RUS and (1+1)EP  are discussed separately.

\subsection{Estimation of Probability to Hit the Promising Region}
\subsubsection{$n=1$}
For this case, we know $x=\sqrt{2M+1-C}>0$.
\begin{itemize}
  \item If $M\le C\le M+1$, it holds that
  \begin{align}
  &P_{cht}^{(1)}(C)=\frac{1}{\sqrt{2\pi}\sigma}\int_{-\sqrt{M}}^{\sqrt{M}}e^{-\frac{(y-\sqrt{2M+1-C})^2}{2\sigma^2}}dy\nonumber\\
  =&\textstyle\Phi\left(\frac{\sqrt{2M+1-C}+\sqrt{M}}{\sigma}\right)-\Phi\left(\frac{\sqrt{2M+1-C}-\sqrt{M}}{\sigma}\right);\label{E3}
  \end{align}
  \item if $0< C<M$,
  \begin{align}
  &P_{cht}^{(1)}(C)=\frac{1}{\sqrt{2\pi}\sigma}\int_{-\sqrt{C}}^{\sqrt{C}}e^{-\frac{(y-\sqrt{2M+1-C})^2}{2\sigma^2}}dy\nonumber\\
  =&\textstyle\Phi\left(\frac{\sqrt{2M+1-C}+\sqrt{C}}{\sigma}\right)-\Phi\left(\frac{\sqrt{2M+1-C}-\sqrt{C}}{\sigma}\right);\label{E4}
  \end{align}
\end{itemize}

Denote $F(\sigma)=\Phi\left(\frac{b}{\sigma}\right)-\Phi\left(\frac{a}{\sigma}\right)$, where $b>a>0$. Then,
\begin{align}%\label{}
  &\frac{d}{d\sigma}F(\sigma)=\frac{1}{\sqrt{2\pi}}\left[e^{-\frac{b^2}{2\sigma^2}}(-\frac{b}{\sigma^2})-e^{-\frac{a^2}{2\sigma^2}}(-\frac{a}{\sigma^2})\right]\nonumber\\
  =&\frac{1}{\sqrt{2\pi}}e^{-\frac{b^2}{2\sigma^2}}(-\frac{b}{\sigma^2})\left[1-e^{-\frac{a^2-b^2}{2\sigma^2}}\frac{a}{b}\right],
\end{align}
which equals zero when $\sigma=\sqrt{\frac{b^2-a^2}{2(\ln b-\ln a)}}$. Then, Cauchy's Mean Value Theorem \cite{thomas2010thomas} implies that $\exists\xi\in(a,b)$ such that
$$\sigma=\sqrt{\frac{2\xi}{2/\xi}}=\xi.$$
 That is to say, to make the probability to hit the promising region as great as possible, $\sigma$ should be endowed with an appropriate value between the upper bound and the lower bound in (\ref{E3}) and (\ref{E4}). Because the results for cases $M\le C\le M+1$ and $0< C< M$ are similar, in the following we only consider the case that $0< C< M$.

\subsubsection{(1+1)RUS} For the $n-D$ case, let the present solution be $\mathbf x=(x_1,\dots,x_n)$ satisfying $\sum_{i=1}^nx_i^2=2M+1-C$. Without loss of generality, suppose that $x_i\ge 0,i=1,\dots,n$. When $x_i$ is selected with probability $\frac{1}{n}$ to be mutated, other $n-1$ variables keep unchanged. Then, a better solution $\mathbf y=(x_1,\dots,x_{i-1},y_i,x_{i+1},\dots,x_n)$ is generated if and only if $y_i^2+\sum_{j\neq i}x_j^2<C$, which is equivalent to $$y_i^2<2(C-M)+1+x_i^2.$$
It does not hold for any $y_i$ when $C$ is sufficiently small. For this case, $\mathbf x$ is located far from the coordination origin, and  then we have $2(C-M)+1+x_i^2<0$. That is to say, (1+1)RUS cannot guarantee convergence to the global optimal solution of (\ref{Cheat}).

\subsubsection{(1+1)EP}For an solution $\mathbf{x}=(x_1,\dots,x_n)$ with $f_{cht}(\mathbf x)=C$, we hanve $
\sum_{i=1}^{n}x_i^2=2M+1-C$, and the promising region $\mathcal{G}_{cht}(C)$ is a sphere centered at the origin with a radius $\sqrt{C}$.
Attributed to the symmetry of the probability dense function (CDF) of Gaussian mutation, we can suppose $\mathbf{x}=(\sqrt{2M+1-C},0,\dots,0)$ without loss of generality. Then
\begin{align}
  &P_{cht}^E(C)=\Pr\{\mathbf y\in\mathcal{G}(C)|\|\mathbf{x}\|^2=2M+1-C\}\nonumber\\
  =&\frac{1}{(\sqrt{2\pi}\sigma)^{n}}\int_{\mathbf y\in\mathcal{G}_U}e^{\frac{(y_1-\sqrt{2M+1-C})^2+\sum_{i=2}^ny_i^2}{2\sigma^2}}dy_1\dots dy_n\nonumber\\
  =&\frac{1}{(\sqrt{2\pi}\sigma)^{n}}\int_{-\sqrt{C}}^{\sqrt{C}}e^{-\frac{(y_1-\sqrt{2M+1-C})^2}{2\sigma^2}}dy_1\nonumber\\
  &\int_{\sum_{i=2}^ny_i^2\le\sqrt{C-y_1^2}}e^{-\frac{\sum_{i=2}^ny_i^2}{2\sigma^2}}dy_2\dots dy_n\nonumber\\
  \le&\frac{1}{(\sqrt{2\pi})^{n}}\int_{-\sqrt{C}}^{\sqrt{C}}e^{-\frac{(y_1-\sqrt{2M+1-C})^2}{2\sigma^2}}dy_1 \left(\int_{-\sqrt{C}}^{\sqrt{C}}e^{-\frac{y^2}{2\sigma^2}}dy\right)^{n-1}\nonumber\\
  =&\textstyle\left(\Phi\left(\frac{\sqrt{2M+1-C}+\sqrt{C}}{\sigma}\right)- \Phi\left(\frac{\sqrt{2M+1-C}-\sqrt{C}}{\sigma}\right)\right)\nonumber\\
  &\textstyle\left(\Phi\left(\frac{\sqrt{C}}{\sigma}\right)-\Phi\left(-\frac{\sqrt{C}}{\sigma}\right)\right)^{n-1}. \label{ProbSphnDU1}
\end{align}

A lower bound of $P_E(C)$ can be obtained by supposing $\mathbf{x}=(x,\dots,x)=(\sqrt{\frac{2M+1-C}{n}},\dots,\sqrt{\frac{2M+1-C}{n}})$. Then, we know that
$\mathcal{G}_{cht}(C)\supset \mathcal{G}_B$.
By (\ref{I_2}) we know that
\begin{align}\label{ProbSphnDU2}
  &P_{cht}^E(C)=\Pr\{\mathbf y\in\mathcal{G}(C)|\mathbf x\}\ge \Pr\{\mathbf y\in\mathcal{G}_B|\mathbf x\}\nonumber\\
  = &\textstyle\left(\Phi\left(\frac{\sqrt{2M+1-C}+\sqrt{C}}{\sigma\sqrt{n}}\right)-\Phi\left(\frac{\sqrt{2M+1-C}-\sqrt{C}}{\sigma\sqrt{n}}\right)\right)^n.
\end{align}
From (\ref{ProbSphnDU1}) and (\ref{ProbSphnDU2}) we conclude that
\begin{align}%\label{}
  P_{cht}^E(C)=\mathcal{O}(a^{n}),\quad a\in(0,1).
\end{align}
%Denote $x=\lfloor x\rfloor+\dot{x}$, where $\lfloor x\rfloor$ represents the integer of $x$, and $\dot{x}$ the fractional part. Just as illustrated in Fig. \ref{RAS},
%\begin{align*}
%  P(C)&=\int_{\mathbf y\in\bigcup\limits_{k=1}^{n}\mathcal{G}_k(C)}d\mathbf P\ge\int_{\mathbf y\in\mathcal{G}_1(C)}d\mathbf P\\
%  &\ge\frac{1}{\sqrt{2\pi}\sigma}\int_{x-1-2\dot{x}}^{x-1}e^{-\frac{(y-x)^2}{2\sigma^2}}dy\\
%  &=\frac{1}{\sqrt{2\pi}}\int_{-(1+2\dot{x})/\sigma}^{-1/\sigma}e^{-\frac{z^2}{2}}dz
%\end{align*}
%\begin{figure}[htbp]
%\centerline{\includegraphics[height=5cm]{Rastrigin.png}}
%\caption{Illustration of the promising region for 1-D Rastrigin's function.}
%\label{RAS}
%\end{figure}

\subsection{Estimation of the  One-step Conditional Improvement Rate}
\subsubsection{$n=1$} The elitist selection contributes to one-step expected  improvement of fitness
\begin{align}\label{FitSph1D1}
  &\mathbb{E}[f(x)-f(y)|x]\nonumber\\
  =&\frac{1}{\sqrt{2\pi}\sigma}\int_{-\sqrt{C}}^{\sqrt{C}}(C-y^2)e^{-\frac{(y-x)^2}{2\sigma^2}}dy\nonumber\\
  =& \textstyle C\left(\Phi\left(\frac{x+\sqrt{C}}{\sigma}\right)-\Phi\left(\frac{x-\sqrt{C}}{\sigma}\right)\right)\nonumber\\
  & -\frac{1}{\sqrt{2\pi}}\int_{-\frac{x+\sqrt{C}}{\sigma}}^{\frac{\sqrt{C}-x}{\sigma}}(\sigma y+x)^2e^{-\frac{y^2}{2}}dy\nonumber\\
  =&\textstyle C\left(\Phi\left(\frac{x+\sqrt{C}}{\sigma}\right)-\Phi\left(\frac{x-\sqrt{C}}{\sigma}\right)\right)-\textstyle\left[\sigma^2I_1(-\frac{x+\sqrt{C}}{\sigma},\frac{\sqrt{C}-x}{\sigma})\right.\nonumber\\
  +&\textstyle\left.x^2I_2(-\frac{x+\sqrt{C}}{\sigma},\frac{\sqrt{C}-x}{\sigma})+2\sigma xI_3(-\frac{x+\sqrt{C}}{\sigma},\frac{\sqrt{C}-x}{\sigma})\right]
\end{align}
Then, (\ref{FitSph1D1}), (\ref{I_1}), (\ref{I_2}) and (\ref{I_3}) imply that
\begin{align*}
  &\mathbb{E}\left[f(x)-f(x)|x\right]\\
   =&\textstyle(C-\sigma^2-x^2)\left[\Phi(\frac{x+\sqrt{C}}{\sigma})-\Phi(\frac{x-\sqrt{C}}{\sigma})\right]\\
  &+\textstyle\frac{\sigma(\sqrt{C}+x)}{\sqrt{2\pi}}e^{-\frac{(\sqrt{C}-x)^2}{\sigma^2}}+\frac{\sigma(\sqrt{C}-x)}{\sqrt{2\pi}}e^{-\frac{(\sqrt{C}+x)^2}{\sigma^2}},
\end{align*}
and the conditional improvement rate is
\begin{align}
  & IR_{cht}^{(1)}(C)=\mathbb{E}\left[{f(x)-f(x)}|x\right]/C\nonumber\\
   =&\textstyle\left(1-\frac{\sigma^2+x^2}{C}\right)\left[\Phi(\frac{x+\sqrt{C}}{\sigma})-\Phi(\frac{x-\sqrt{C}}{\sigma})\right]\nonumber\\
  &+\textstyle\frac{\sigma(\sqrt{C}+x)}{\sqrt{2\pi}C}e^{-\frac{(\sqrt{C}-x)^2}{\sigma^2}}+\frac{\sigma(\sqrt{C}-x)}{\sqrt{2\pi}C}e^{-\frac{(\sqrt{C}+x)^2}{\sigma^2}},\label{OneStepImpRate11}
\end{align}
where $x=\sqrt{2M+1-C}$.

\subsubsection{(1+1)EP}

Similarly, the one-step improvement of fitness is bounded from above by
\begin{align}\label{OneStepImpRaten11}
  &IR_{cht}^E(C)\le P_{cht}^E(C)\nonumber\\
 = &\textstyle \left(\Phi\left(\frac{\sqrt{2M+1-C}+\sqrt{C}}{\sigma}\right)- \Phi\left(\frac{\sqrt{2M+1-C}-\sqrt{C}}{\sigma}\right)\right)\nonumber\\
  &\textstyle\left(\Phi\left(\frac{\sqrt{C}}{\sigma}\right)-\Phi\left(-\frac{\sqrt{C}}{\sigma}\right)\right)^{n-1}. \end{align}

To obtain a lower bound of $IR_{cht}^E(C)$, we suppose
$$\mathbf{x}=(x,\dots,x)=(\sqrt{\frac{2M+1-C}{n}},\dots,\sqrt{\frac{2M+1-C}{n}}).$$
Then,
\begin{align*}%\label{}
  &\mathbb{E}[f(\mathbf{x})-f(\mathbf{y})|\|\mathbf{x}\|^2=C]\\
  &p=\int_{\mathcal G_{cht}(C)}(C-\sum_{i=1}^ny_i^2)d\mathbf{P}\ge \int_{\mathcal G_B}(C-\sum_{i=1}^ny_i^2)d\mathbf{P}\nonumber\\
  &=\textstyle C\left(\Phi\left(\frac{\sqrt{2M+1-C}+\sqrt{C}}{\sigma\sqrt{n}}\right)-\Phi\left(\frac{\sqrt{2M+1-C}-\sqrt{C}}{\sigma\sqrt{n}}\right)\right)^n\nonumber\\
  &-\textstyle n IR_{cht}^{(1)}p(C)\left(\Phi\left(\frac{\sqrt{2M+1-C}+\sqrt{C}}{\sigma\sqrt{n}}\right)-\Phi\left(\frac{\sqrt{2M+1-C}-\sqrt{C}}{\sigma\sqrt{n}}\right)\right)^{n-1}\nonumber\\
  &=\textstyle\left(\Phi\left(\frac{\sqrt{2M+1-C}+\sqrt{C}}{\sigma\sqrt{n}}\right)-\Phi\left(\frac{\sqrt{2M+1-C}-\sqrt{C}}{\sigma\sqrt{n}}\right)\right)^{n-1}\nonumber\\
  &\textstyle\left[C\left(\Phi\left(\frac{\sqrt{2M+1-C}+\sqrt{C}}{\sigma\sqrt{n}}\right)-\Phi\left(\frac{\sqrt{2M+1-C}-\sqrt{C}}{\sigma\sqrt{n}}\right)\right)-nIR(C)\right],
\end{align*}
and
\begin{align}
  &IR_{cht}^E(C) \ge\textstyle\left(\Phi\left(\frac{\sqrt{2M+1-C}+\sqrt{C}}{\sigma\sqrt{n}}\right)-\Phi\left(\frac{\sqrt{2M+1-C}-\sqrt{C}}{\sigma\sqrt{n}}\right)\right)^{n-1}\nonumber\\
  &\textstyle\left[\left(\Phi\left(\frac{\sqrt{2M+1-C}+\sqrt{C}}{\sigma\sqrt{n}}\right)-\Phi\left(\frac{\sqrt{2M+1-C}-\sqrt{C}}{\sigma\sqrt{n}}\right)\right)-\frac{nIR_{cht}^E(C)}{C}\right],\label{OneStepImpRaten111}
\end{align}
where $IR_{cht}^{(1)}(C)$ is confirmed by (\ref{OneStepImpRate11}). Combining (\ref{OneStepImpRaten11}) and (\ref{OneStepImpRaten111}) we know that
\begin{align}%\label{}
  IR_{cht}^E(C)=\Theta(a^n),\quad a\in(0,1).
\end{align}

\section{Conclusion}\label{ConSec}
This paper propose to evaluate exploitation/exploration of EAs by either the probability to hit the promising region or the one-step improvement rate, instead of taking them as two conflicting items. By case studies, it is demonstrated that both exploitation and exploration ability of EAs, evaluated by the probability or the improvement rate, degenerate with the problem dimension $n$.  It is not surprising that the search algorithm (1+1)RUS performs exploitation better than (1+1)EP, but the global convergence to optima of the cheating problem can only be achieved by the (1+1)EP.

The results on connections between exploitation (exploration) and $\sigma$ demonstrate significant difference between exploration and exploitation. For exploration, estimations of both probability and improvement rate show that best performance is achieved by setting $\sigma$ to an appropriate value greater than $0$; however, exploitation analysis indicate that the success probability to hit the promising region increase as $\sigma$ decrease, which contradicts to the fact that the one-step improvement rate achieve the maximum value at some $\sigma^*>0$. Our future work will focus on discover quantitative relation between the exploitation/exploration metrics and consecutive convergence rate of EAs, and then, try to get a parameter setting strategy for optimization of EA's performance.

\end{document}